\documentclass[conference,compsoc]{ieeeconf}  

\IEEEoverridecommandlockouts                              

\usepackage{cite}
\usepackage{xcolor,soul}
\usepackage{tcolorbox} 
\soulregister\cite7
\soulregister\citep7
\soulregister\citet7
\soulregister\ref7
\soulregister\pageref7

\usepackage{amssymb}
\usepackage{bm}
\usepackage{flushend}

\usepackage{multirow}
\usepackage{soul}
\usepackage{svg}
\usepackage{float}

\usepackage{hyperref}

\usepackage{cuted}
\usepackage{capt-of}
\usepackage{caption}
\usepackage{amsmath}

\usepackage{blindtext}
\usepackage{tabularx}

\usepackage{amssymb}
\usepackage{bm}
\usepackage{balance}
\usepackage{titlesec}
\usepackage{flushend} 

\usepackage[switch,pagewise]{lineno}
\usepackage{hyperref}
\hypersetup{
    colorlinks=true,
    linkcolor=blue,
    filecolor=blue,      
    urlcolor=blue,
    citecolor=cyan,
}

\title{\bf HelmetPoser: A Helmet-Mounted IMU Dataset for Data-Driven Estimation of Human Head Motion in Diverse Conditions}

\author{Jianping Li\textsuperscript{\textdagger}, Qiutong Leng\textsuperscript{\textdagger}, Jinxin Liu, Xinhang Xu, Tongxin Jin, \\Muqing Cao, Thien-Minh Nguyen, Shenghai Yuan, Kun Cao, Lihua Xie,~\IEEEmembership{Fellow,~IEEE}}

\begin{document}
\maketitle


\renewcommand{\thefootnote}{}
\footnotetext{This research is supported by China-Singapore International Joint Research Institute (CSIJRI) Development Plan for the Technology Application Center (TAC) and the National Research Foundation, Singapore, under its Medium-Sized Center for Advanced Robotics Technology Innovation (CARTIN). 

All authors are with China-Singapore International Joint Research Institute (CSIJRI) and School of Electrical and Electronic Engineering, Nanyang Technological University, Singapore 639798, 50 Nanyang Avenue. Kun Cao is with the Department of Control Science and Engineering, College of Electronics and Information Engineering, and Shanghai Research Institute for Intelligent Autonomous Systems, Tongji University, Shanghai, China. (E-mail: jianping.li@ntu.edu.sg, leng0039@e.ntu.edu.sg, jinxin.liu@ntu.edu.sg, xinhang.xu@ntu.edu.sg, tongxing.jin@ntu.edu.sg, mqcao@ntu.edu.sg, thienminh.nguyen@ntu.edu.sg, shyuan@ntu.edu.sg, caokun@tongji.edu.cn, elhxie@ntu.edu.sg) (Jianping Li and Qiutong Leng are co-first authors.)}


\begin{abstract}
Helmet-mounted wearable positioning systems are crucial for enhancing safety and facilitating coordination in industrial, construction, and emergency rescue environments. These systems, including LiDAR-Inertial Odometry (LIO) and Visual-Inertial Odometry (VIO), often face challenges in localization due to adverse environmental conditions such as dust, smoke, and limited visual features. To address these limitations, we propose a novel head-mounted Inertial Measurement Unit (IMU) dataset with ground truth, aimed at advancing data-driven IMU pose estimation. Our dataset captures human head motion patterns using a helmet-mounted system, with data from ten participants performing various activities. We explore the application of neural networks, specifically Long Short-Term Memory (LSTM) and Transformer networks, to correct IMU biases and improve localization accuracy. Additionally, we evaluate the performance of these methods across different IMU data window dimensions, motion patterns, and sensor types. We release a publicly available dataset, demonstrate the feasibility of advanced neural network approaches for helmet-based localization, and provide evaluation metrics to establish a baseline for future studies in this field. Data and code can be found at \url{https://lqiutong.github.io/HelmetPoser.github.io/}.
    
\end{abstract}

\section{Introduction }

\begin{figure}[]
    \centering
    \includegraphics[width=\linewidth]{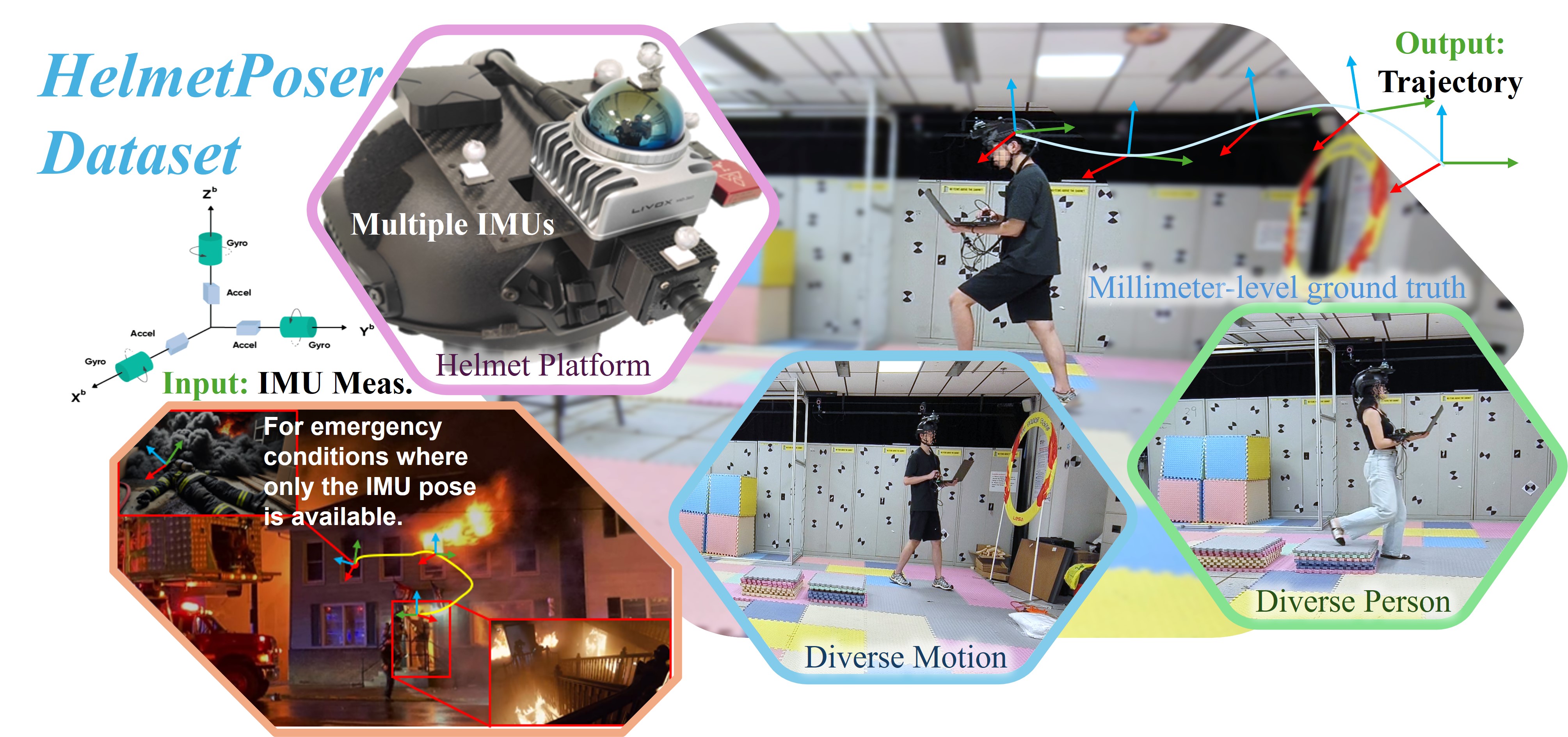} 
    \caption{HelmetPoser dataset for data-driven pose estimation of human head motion using IMU, especially for emergency conditions. The dataset is collected using the helmet platform with multiple IMUs with diverse motions and diverse persons. The millimeter-level ground truth is obtained using the VICON system.}
    \label{fig:abstract}
    \vspace{-0.3cm}
\end{figure}

Helmet-mounted wearable positioning systems are essential for enhancing safety and facilitating collaboration between agents across multiple fields \cite{hillers2004terebes,hoffman2006virtual,li2024hcto}. In industrial and construction settings, these devices are instrumental in tracking and managing workers. By delivering accurate location data, they enable managers to monitor workers in real-time, ensuring they stay within designated safe zones and thereby minimizing the risk of accidents \cite{campero2020smart}. In emergency rescue situations, helmet-mounted positioning devices are indispensable tools for rescue teams. They deliver real-time location data of rescuers, enabling command centers to coordinate operations more effectively, enhance efficiency, and reduce response times \cite{lee2019efficient}. However, in challenging industrial construction and rescue environments, integrated sensor systems in helmets, such as LiDAR-Inertial Odometry (LIO) \cite{xu2022fast} and Visual-Inertial Odometry (VIO) \cite{qin2018vins}, may still struggle with localization due to adverse environmental conditions \cite{li2023whu,li2023real,nguyen2024eigen}. The failure of these multi-sensor Simultaneous Localization and Mapping (SLAM) algorithms can be attributed to several factors:

Firstly, LiDAR sensors rely on the reflection of light signals to measure distances. In environments with heavy dust, smoke, or moisture, the light signals can be scattered or absorbed, leading to inaccurate environmental data acquisition by LiDAR. Additionally, in narrow spaces or areas with minimal environmental features, the reflected signals may be insufficient, causing LiDAR to function ineffectively. For example, in tunnels or mines, the performance of LiDAR sensors can significantly degrade.

Secondly, visual sensors depend on spectral features \cite{mur2015orb} captured by cameras for localization. When there is insufficient ambient light, dynamic blur, or visual occlusion, cameras may fail to capture enough feature points, resulting in visual tracking failures. For instance, in dark environments or areas filled with smoke, the effectiveness of VIO systems can be markedly reduced. This issue is exacerbated in rescue scenarios where rapidly changing environments and high-intensity dynamic movements increase the difficulty of visual tracking.

Different from the above-mentioned sensors, IMUs estimate motion states by measuring acceleration and angular velocity, providing the advantage of not relying on external environmental conditions. However, IMUs are subject to error accumulation, where small measurement errors gradually accumulate over time, leading to increasing inaccuracies in position and orientation estimates. This phenomenon, known as drift, makes long-term reliance on IMUs for localization problematic.

In recent years, various data-driven methods have been proposed to address the error accumulation problem in Inertial Measurement Units (IMUs), such as correcting IMU biases \cite{buchanan2022deep} or motion drift \cite{liu2020tlio,han2019deepvio} through deep learning techniques. However, most current research relies on data sets collected on specific platforms, such as vehicles \cite{liao2022kitti}, robotic platforms \cite{mcdviral2024,,li2019nrli,li2024graph}, and the human body \cite{yan2019ronin}. There are still few datasets that capture the high-frequency motion patterns of the human head, as experienced with helmet usage. The motion patterns of the human head are highly non-linear and uncertain \cite{li2023whu}, complicating the potential of data-driven IMU motion estimation. This gap presents a significant challenge for achieving high-precision localization systems using the data-driven IMU state estimation on the helmets. 

Thus, we proposed a novel head-mounted IMU dataset with ground truth to push the limits of data-driven IMU pose estimation. The contributions of this research are primarily threefold:

(1) We introduce a publicly available dataset that captures human head motion patterns using a helmet-mounted system. This dataset includes motion data from five males and five females, covering a range of daily activities such as walking, running, and stair climbing. It can be used to advance and validate various localization methods.
    
(2) We demonstrated the feasibility of using neural networks to learn IMU biases on helmet-mounted localization devices. Specifically, we employed LSTM and Transformer networks and tested them on the dataset to evaluate their effectiveness.

(3) We assessed the performance of the proposed method by varying the dimensions of the IMU data window, different motion patterns, and various IMU sensors. We provided evaluation metrics to establish a baseline for helmet-based localization devices, thus supporting future research in this domain.

\section{Methodology} \label{section_method}

To collect the helmet-mounted IMU dataset with ground truth, we first set up the hardware of the helmet system \ref{sensor_setup}. Then, we use the IMU pre-integration to derive the IMU bias according to the ground truth trajectory obtained from the VICON system \ref{preintegration_IMU_bias}. Finally, different networks are detailed for IMU bias prediction \ref{networks_structures_IMU}.

\subsection{Sensor Setup of the Helmet Positioning System} \label{sensor_setup}
Our helmet-based localization system is equipped with a Livox Mid 360 LiDAR sensor, which includes a built-in low-cost IMU, the ICM40609 \cite{ICM40609}. Additionally, the system features an industry-level IMU, VN-100 \cite{vectornav}, widely used in industrial and military applications. The helmet is also fitted with markers to support real-time localization tracking using the VICON system. This combination of sensors ensures that the system is well-suited for both everyday use and industrial applications.

\begin{figure}[ht]
    \centering
    \includegraphics[width=\linewidth]{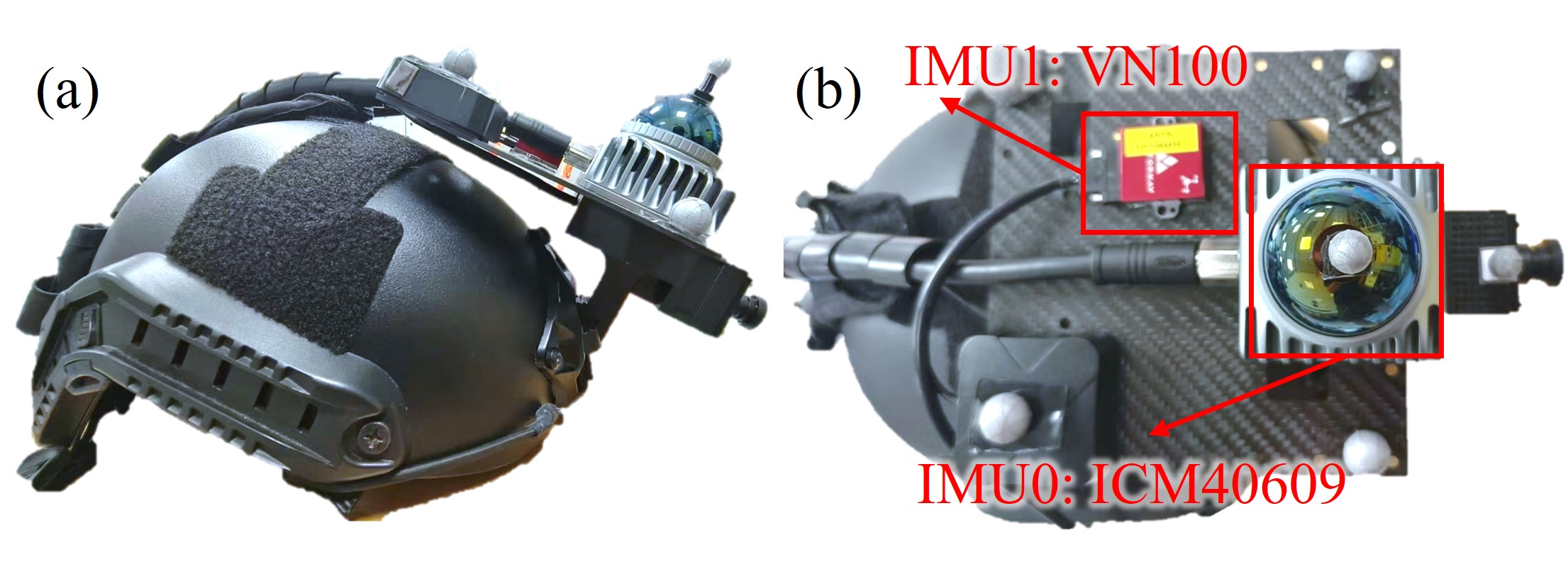} 
    \caption{Sensor setup for the helmet system. (a) Side view; (b) Top view of the helmet including two IMUs at different levels.}
    \label{fig:hardware-setup}
    \vspace{-0.3cm}
\end{figure}

Before calculating the IMU biases, it is necessary to calibrate the helmet coordinate system with the IMU coordinate system. In this process, we need to obtain the rotation matrix from the helmet coordinate system and then the IMU coordinate system. To achieve this, we employ a hand-eye calibration algorithm. For the rotational component of hand-eye calibration, we solve the orthogonal Procrustes problem to find the optimal calibration rotation matrix \( R \) \cite{zhao2011hand}. We obtain the sets of rotation matrices for the helmet and IMU, \( \{A_i\} \) and \( \{B_i\} \), respectively. We seek to find an optimal rotation matrix \( R \) that minimizes the following objective function:

\begin{align}
\text{arg \,min}_R \sum_{i=1}^{n} \| R A_i - B_i \|_F^2
\label{eq:formula1}
\end{align}

where \( \| \cdot \|_F \) denotes the Frobenius norm, which measures the difference between two matrices. The solution involves first computing the matrix \( M \) as the sum of the products of each \( B_i \) and the transpose of \( A_i \). Then, Singular Value Decomposition (SVD) is performed on \( M \) to obtain matrices \( U \) and \( V \). Finally, the optimal rotation matrix \( R \) is obtained by multiplying \( U \) and \( V \).

    


\subsection{Pre-integration for Ground Truth IMU Bias Estimation}\label{preintegration_IMU_bias}



IMU drift primarily results from the gradual accumulation of bias, making the reduction of this bias essential for improving localization accuracy. To determine GT IMU biases, we integrate the raw IMU measurements according to the pre-integration \cite{qin2018vins} rules and compare the integrated values with the GT poses to calculate the biases. Now, we derive the equations for the GT bias calculation.

\subsubsection{IMU Pre-integration} 
The raw gyroscope and accelerometer measurements from the IMU, denoted as $\hat{\omega}$ and $\hat{a}$, are first given by \cite{shin2004unscented}:

\begin{align}
\hat{a}_t &= a_t + \mathbf{b}_{a_t} + \mathbf{R}_w^t \mathbf{g}^w + \mathbf{n}_a, \\
\hat{\omega}_t &= \omega_t + \mathbf{b}_{w_t} + \mathbf{n}_w.
\label{eq:formula23}
\end{align} These measurements are subject to acceleration bias ($\mathbf{b}_a$), gyroscope bias ($\mathbf{b}_w$), and additional noise. By transforming the reference frame from the global coordinate system to a local coordinate system ($b_k$), we focus solely on pre-integrating components related to linear acceleration and angular velocity. Considering that an individual typically takes 0.5 seconds to complete a basic action, such as walking, running, or climbing stairs, we set the time interval between consecutive time points, i.e., between $b_k$ and $b_{k+1}$, to 0.5 seconds. In IMU pre-integration, $\alpha_{b_{k+1}}^{b_{k}}$ represents position integration, $\beta_{b_{k+1}}^{b_{k}}$ represents velocity integration, and $\gamma_{b_{k+1}}^{b_{k}}$ represents orientation integration. In practical scenarios, IMU data is discrete. It is important to note that initially, $\alpha_{b_{k}}^{b_k}$ and $\beta_{b_{k}}^{b_k}$ are set to 0, and $\gamma_{b_{k}}^{b_k}$ is the identity quaternion. The additive noise terms $\mathbf{n}_a$ and $\mathbf{n}_w$ are unknown and are treated as zero during the implementation. Consequently, we can derive the discrete expressions for IMU pre-integration, resulting in the estimated values of the pre-integration terms, denoted by $\hat{(\cdot)}$ as follows.
\begin{align}
\hat{\alpha}_{i+1}^{b_k} &= \hat{\alpha}_i^{b_k} + \hat{\beta}_i^{b_k} \delta t + \frac{1}{2} \mathbf{R}(\hat{\gamma}_i^{b_k})(\hat{\mathbf{a}}_i - \mathbf{b}_{a_i}) \delta t^2 \label{eq4}, \\
\hat{\beta}_{i+1}^{b_k} &= \hat{\beta}_i^{b_k} + \mathbf{R}(\hat{\gamma}_i^{b_k})(\hat{\mathbf{a}}_i - \mathbf{b}_{a_i}) \delta t \label{eq5}, \\
\hat{\gamma}_{i+1}^{b_k} &= \hat{\gamma}_i^{b_k} \otimes \left[ \frac{1}{2} (\hat{\boldsymbol{\omega}}_i - \mathbf{b}_{w_i}) \delta t \right] \label{eq6}.
\end{align} The index $i$ represents a discrete moment corresponding to an IMU measurement within the interval $[t_k, t_{k+1}]$. The time interval $\delta t$ represents the period between two successive IMU measurements $i$ and $i+1$. Using equations (\ref{eq4}) to (\ref{eq6}), we can compute the IMU pre-integration values.

The continuous-time linearized dynamics of error terms of (\ref{eq4}) - (\ref{eq6}) are derived as follows.
\begin{equation}
\small
\begin{aligned}
\begin{bmatrix}
\delta \dot{\alpha}_t^{b_k} \\
\delta \dot{\beta}_t^{b_k} \\
\delta \dot{\boldsymbol{\theta}}_t^{b_k} \\
\delta \dot{\mathbf{b}}_{a_t} \\
\delta \dot{\mathbf{b}}_{w_t}
\end{bmatrix}
&= 
\begin{bmatrix}
0 & \mathbf{I} & 0 & 0 & 0 \\
0 & 0 & -\mathbf{R}_t^{b_k} [\hat{\mathbf{a}}_t - \mathbf{b}_{a_t}]_\times & -\mathbf{R}_t^{b_k} & 0 \\
0 & 0 & -[\hat{\boldsymbol{\omega}}_t - \mathbf{b}_{w_t}]_\times & 0 & -\mathbf{I} \\
0 & 0 & 0 & 0 & 0 \\
0 & 0 & 0 & 0 & 0
\end{bmatrix}
\begin{bmatrix}
\Delta\hat{\alpha}_{b_k+1}^{b_k} \\
\delta \beta_t^{b_k} \\
\delta \boldsymbol{\theta}_t^{b_k} \\
\delta \mathbf{b}_{a_t} \\
\delta \mathbf{b}_{w_t}
\end{bmatrix} \\
&\quad +
\begin{bmatrix}
0 & 0 & 0 & 0 \\
-\mathbf{R}_t^{b_k} & 0 & 0 & 0 \\
0 & -\mathbf{I} & 0 & 0 \\
0 & 0 & \mathbf{I} & 0 \\
0 & 0 & 0 & \mathbf{I}
\end{bmatrix}
\begin{bmatrix}
\mathbf{n}_a \\
\mathbf{n}_w \\
\mathbf{n}_{b_{a_t}} \\
\mathbf{n}_{b_{w_t}}
\end{bmatrix} \\
&= \mathbf{F}_t \delta \mathbf{z}_t^{b_k} + \mathbf{G}_t \mathbf{n}_t,
\end{aligned}
\label{eq7}
\end{equation}
The first-order Jacobian matrix $\mathbf{J}_{b_k+1}$ can also be computed recursively, starting with the initial Jacobian $\mathbf{J}_{b_k} = \mathbf{I}$.

\begin{equation}
\mathbf{J}_{t+\delta t} = (\mathbf{I} + \mathbf{F}_t \delta t) \mathbf{J}_t, \quad t \in [k, k+1].
\label{eq8}
\end{equation}

\subsubsection{GT IMU Bias Calculation} 
We are able to obtain the nominal values of human motion between $b_k$ and $b_{k+1}$ using the VICON system:
\begin{equation}
\small
\begin{bmatrix}
\alpha_{b_k+1}^{b_k} \\
\beta_{b_k+1}^{b_k} \\
\gamma_{b_k+1}^{b_k} \\
0 \\
0
\end{bmatrix}
=
\begin{bmatrix}
\mathbf{R}_{w}^{b_k} (\mathbf{p}_{b_k+1}^w - \mathbf{p}_{b_k}^w + \frac{1}{2} \mathbf{g}^w \Delta t_k^2 - \mathbf{v}_{b_k}^w \Delta t_k) \\
\mathbf{R}_{w}^{b_k} (\mathbf{v}_{b_k+1}^w + \mathbf{g}^w \Delta t_k - \mathbf{v}_{b_k}^w) \\
\mathbf{q}_{b_k}^{w^{-1}} \otimes \mathbf{q}_{b_k+1}^w \\
\mathbf{b}_{a_{b_k+1}} - \mathbf{b}_{a_{b_k}} \\
\mathbf{b}_{w_{b_k+1}} - \mathbf{b}_{w_{b_k}}
\end{bmatrix},
\label{eq9}
\end{equation}
where $\alpha_{b_{k+1}}^{b_{k}}$, $\beta_{b_{k+1}}^{b_{k}}$, and $\gamma_{b_{k+1}}^{b_{k}}$ are calculated using the GT poses obtained by the VICON system.
Utilizing the recursive formulation (\ref{eq8}), the first-order approximations of $\alpha_{b_{k+1}}^{b_{k}}$, $\beta_{b_{k+1}}^{b_{k}}$, and $\gamma_{b_{k+1}}^{b_{k}}$ related to biases can be expressed as follows:
\begin{equation}
\begin{aligned}
\alpha_{b_k+1}^{b_k} &\approx \hat{\alpha}_{b_k+1}^{b_k} + \mathbf{J}^{\alpha}_{b_a} \delta \mathbf{b}_{a_k} + \mathbf{J}^{\alpha}_{b_w} \delta \mathbf{b}_{w_k}, \\
\beta_{b_k+1}^{b_k} &\approx \hat{\beta}_{b_k+1}^{b_k} + \mathbf{J}^{\beta}_{b_a} \delta \mathbf{b}_{a_k} + \mathbf{J}^{\beta}_{b_w} \delta \mathbf{b}_{w_k}, \\
\gamma_{b_k+1}^{b_k} &\approx \hat{\gamma}_{b_k+1}^{b_k} \otimes \left[ \frac{1}{2} \mathbf{J}^{\gamma}_{b_w} \delta \mathbf{b}_{w_k} \right].
\end{aligned}
\label{eq10}
\end{equation}
Finally, by solving the Jacobian first-order approximations of $\alpha_{b_{k+1}}^{b_{k}}$, $\beta_{b_{k+1}}^{b_{k}}$, and $\gamma_{b_{k+1}}^{b_{k}}$, we can obtain the IMU bias values at each time step. In (\ref{eq10}), the terms on the right side of the equations with $\hat{(\cdot)}$ represent the pre-integrated IMU values, while the terms on the left side represent the ground truth values calculated by the Vicon system. At each time step (from $b_k$ to $b_{k+1}$), we assume that the biases remain constant.

\subsection{Networks Structures for IMU Bias Prediction}\label{networks_structures_IMU}

After obtaining the GT IMU biases, we use the networks to predict the IMU biases. Here, LSTM \cite{yu2019review} and Transformer \cite{han2022survey} networks were employed to learn and predict the IMU data biases. LSTM is adept at handling time-series data, effectively capturing long-term dependencies through its memory cells, which helps maintain accurate bias estimation over extended periods. On the other hand, Transformers process long time-series data through self-attention mechanisms, capturing global dependencies. This not only enables them to learn the patterns of bias changes but also supports parallel computation, thereby improving training and inference efficiency.

\subsubsection{LSTM}

\begin{figure}[]
    \centering
    \includegraphics[width=\columnwidth]{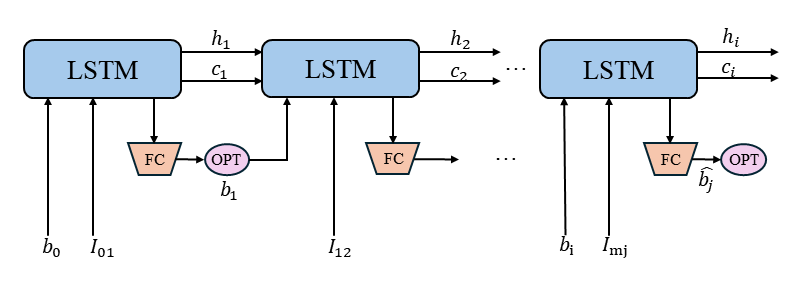} 
    \caption{LSTM architecture. An IMU data input  $\mathcal{I}_{mj}$ of size $w$ (with $m = j - w$) and the previous bias $b_i$ are passed to the LSTM. The hidden state is preserved for the next inference step and the output is passed through a fully connected layer to predict a bias.}
    \label{fig:lstm-architecture}
    \vspace{-0.5cm}
\end{figure}

The input to the LSTM consists of a series of IMU data windows. A single window of dimensions $w$ of IMU measurements $I_{mj}$ (with $m = j - w$) and the previous bias estimate $b_i$, which comes from the IMU pre-integration. These are normalized and then passed to the LSTM with states $h_i$ and $c_i$ which are preserved for the next bias estimate. In this way, the LSTM processes $w$ data of IMU measurements while the memory can observe the bias evolution over time.

We use the original network's suggested configuration: a 2-layer single-direction LSTM with a hidden state dimension of 256. The input dimensions of IMU data is 10 (i.e., $w = 10$) and the history window of inputs is 32. Each window corresponds to a one-time step in the IMU pre-integration (from $b_k$ to $b_{k+1}$). To enable the neural network to learn more comprehensive patterns, we ensured a 50\% data overlap between consecutive inferences during the training process. This means that the LSTM learns from 32 consecutive steps each time and includes the last 16 steps in the next loaded batch. This approach helps the network learn the continuous evolution of IMU bias changes and make better predictions.

\subsubsection{Transformer}

\begin{figure}[ht]
    \centering
    \includegraphics[width=\columnwidth]{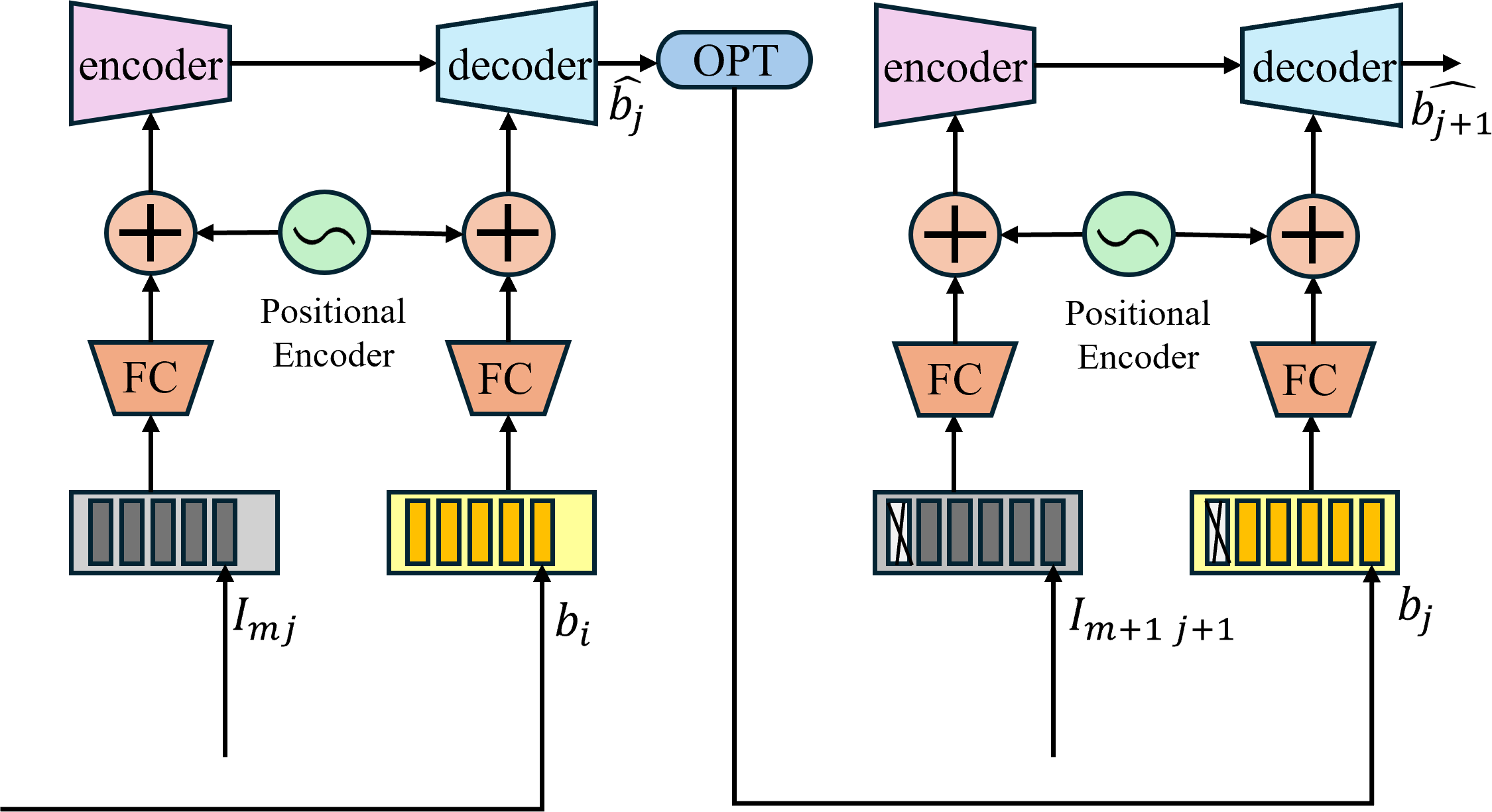}
    \caption{Transformer architecture. A sequence of IMU windows $I_{mj}$(with history $l$) and biases $b_i$ are combined and integrated with positional encoding before being input into the Transformer. }
    \vspace{-0.5cm}
    \label{fig:transformer}
\end{figure}

The Transformer input is a history of $l$ windows of IMU measurements and biases. Similar to the LSTM, biases added to the history come from the IMU pre-integration. A history of information allows the Transformer attention mechanism to recall older information.

The original model suggested an 8-headed Transformer with 2 encoder and decoder layers and an embedding dimension of 512. Unlike the LSTM, the Transformer's capability allows it to input 100 IMU data windows ($l = 100$). Data overlap is not required here.

\subsubsection{Loss Function} 
    
We use the Mean Square Error (MSE) as a loss function:
\begin{equation}
\mathcal{L}(\mathbf{b}, \hat{\mathbf{b}}) = \frac{1}{n} \sum_{k=1}^{n} \| \mathbf{b}_k - \hat{\mathbf{b}}_k \|^2
\label{eq11}
\end{equation}
where two separate instances of the network are trained with (20) for accelerometer and gyroscope biases, respectively.


\begin{figure}[]
    \centering
    \includegraphics[width=\linewidth]{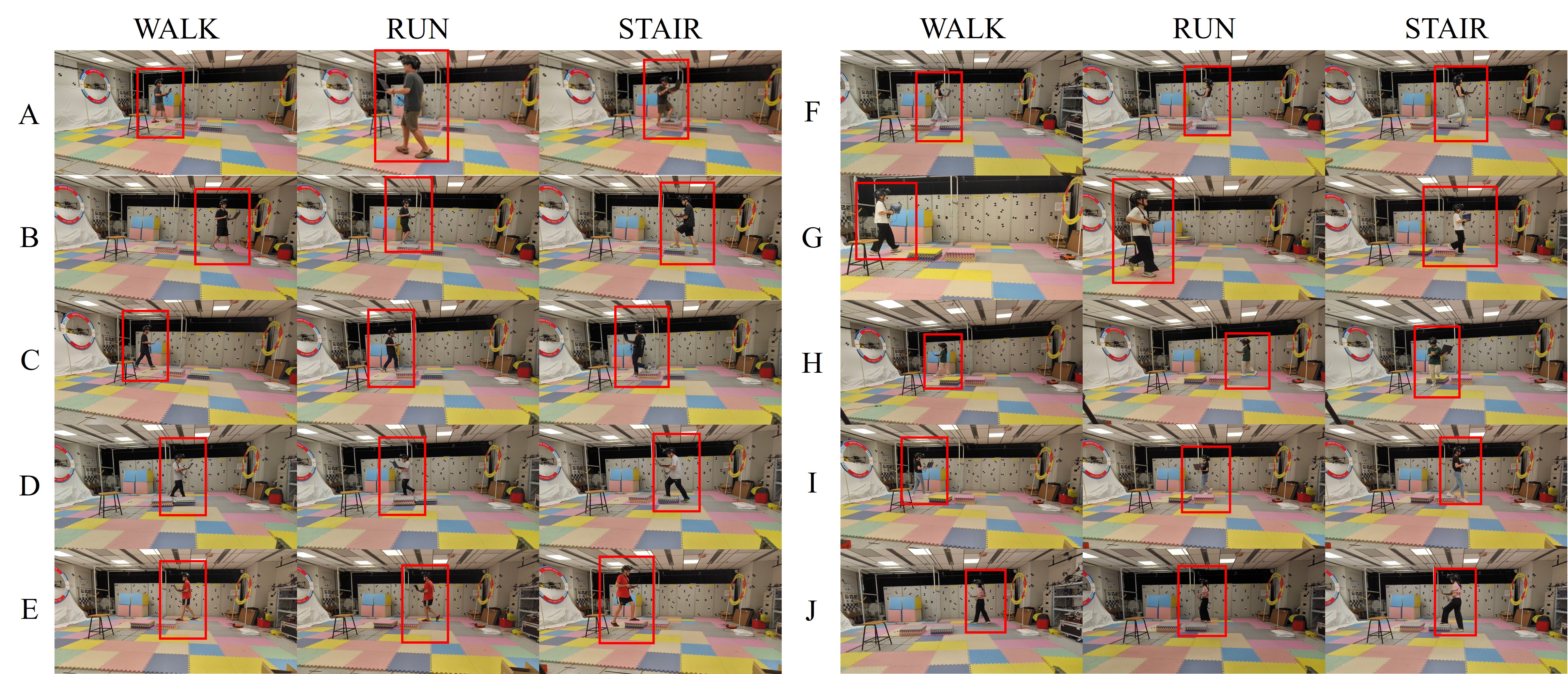}
    \caption{Dataset Information: The dataset was recorded with ten participants, labeled A through J. The first five participants are male, and the remaining five are female. The diagram illustrates three actions performed by each participant, shown from left to right: walking, running, and stair climbing.}
    \label{fig:experiment}
    \vspace{-0.5cm}
\end{figure}

\section{HelmetPoser Dataset} \label{section_exp}

\subsection{Data Collection}

In the experiment, we record data for three primary actions: walking, running, and stair climbing. Walking represents the most common daily movement, running tests the performance of IMU sensors in dynamic environments, and stair climbing introduces vertical height changes and varying gait patterns. Participants wear helmet-based localization devices and record data in a laboratory equipped with a VICON system. Each participant completes three sessions, each lasting approximately five minutes. This dataset provides a comprehensive and reliable foundation for validating our methods.

During the preparation of the dataset, a total of ten participants were involved in data collection. The following provides their basic information:

\begin{table}[]
    \caption{Participant Characteristics\label{tab:sample_data}}
    \centering
    \begin{tabularx}{\columnwidth}{|>{\centering\arraybackslash}X|>{\centering\arraybackslash}X|>{\centering\arraybackslash}X|>{\centering\arraybackslash}X|}\hline\hline
    Name & Height(M) & Weight(KG) & Gender \\ \hline
    A & 1.78 & 66 & Male \\ 
    B & 1.84 & 74 & Male \\ 
    C & 1.75 & 63 & Male \\ 
    D & 1.72 & 66 & Male \\ 
    E & 2.08 & 85 & Male \\ 
    F & 1.65 & 45 & Female \\ 
    G & 1.6 & 42 & Female \\ 
    H & 1.55 & 40 & Female \\ 
    I & 1.7 & 44 & Female \\ 
    J & 1.72 & 51 & Female \\ \hline\hline
    \end{tabularx}
    \label{table1}
    \vspace{-0.5cm}
\end{table}

Our recorded datasets are named using a combination of the participant's identifier and the action number, where the suffix \textbf{w} represents walking, \textbf{r} represents running, and \textbf{c} represents stair climbing. For example, A\_w indicates the walking data of participant A.

\subsection{Data Format}


Our data is recorded in ROSbag format, which is commonly used in ROS for logging sensor data and other related information. The ROSbag file contains data from three topics: \texttt{/livox/imu}, \texttt{/vectorNAV/IMU}, and \texttt{/vicon/helmet}. Specifically, \texttt{/livox/imu} records IMU data from the Livox Mid-360, \texttt{/vectorNAV/IMU} includes acceleration and angular velocity information from the VectorNav IMU, and \texttt{/vicon/helmet} logs the helmet position and orientation of the participants during the experiment under the VICON system.

\begin{table}[]
    \caption{Sensor Topics and Publishing Frequency\label{tab:sensor_topics}}
    \centering
    \begin{tabularx}{\columnwidth}{|>{\centering\arraybackslash}X|>{\centering\arraybackslash}X|>{\centering\arraybackslash}X|}\hline\hline
    Sensor & Topic Name & Publishing Frequency \\ \hline
    Livox Mid 360 & \texttt{/livox/imu} & 200 Hz \\ 
    VectorNav IMU & \texttt{/vectorNAV/IMU} & 200 Hz \\ 
    Vicon Helmet & \texttt{/vicon/helmet} & 50 Hz \\ \hline\hline
    \end{tabularx}
    \label{tb2}
\end{table}

\section{Evaluation and Analysis of Data-driven Head Motion Estimation Networks} \label{section_evaluation}

\subsection{Accuracy Analysis of the Networks}

We select the dataset from participant D as the validation set for training and conducted 200 epochs of training. The study find that when the input sequence length of the IMU data is set to 32, the validation set exhibits a relatively fast convergence, reaching a stable state around the 70th epoch. The loss values of the original LSTM and Transformer models are shown in Table {\ref{tb3}}.

\begin{table}[]
    \caption{Original Model Loss Vulue\label{tab:performance}}
    \centering
    \begin{tabularx}{\columnwidth}{|>{\centering\arraybackslash}X|>{\centering\arraybackslash}X|>{\centering\arraybackslash}X|>{\centering\arraybackslash}X|}\hline\hline
    \multirow{2}{*}{Model} & \multicolumn{3}{c|}{Test Dataset Loss} \\ \cline{2-4}
    & D1 & D2 & D3 \\ \hline
    LSTM & 0.028 & 0.020 & 0.012 \\ \hline
    Transformer & 0.027 & 0.021 & 0.024 \\ \hline\hline
    \end{tabularx}
    \label{tb3}
    \vspace{-0.3cm}
\end{table}

We adopt the input window as suggested by the original text, with the LSTM model using an input window of 32 steps (equivalent to 16 seconds) and the Transformer model using an input sequence length of 100 steps (equivalent to 50 seconds). We calculate the $\Delta{\alpha}$ before and after compensation and use the percentage reduction in $\Delta{\alpha}$ as the performance metric for the model. This performance metric effectively illustrates the improvement in IMU accuracy before and after compensation. The performance metric formula for the compensation is as follows:

\begin{align}
\text{Performance Metric} =
\frac{\Delta{\alpha}(before) - \Delta{\alpha}(after)}{\Delta{\alpha}(before)}
\label{eq12}
\end{align}

The models' performance on test sets D1, D2, and D3 is as follows:

\begin{table}[]
    \caption{Models' Performance Metric\label{tab:performance}}
    \centering
    \begin{tabularx}{\columnwidth}{|>{\centering\arraybackslash}X|>{\centering\arraybackslash}X|>{\centering\arraybackslash}X| >{\centering\arraybackslash}X|}\hline\hline
    \multirow{2}{*}{Model} & \multicolumn{3}{c|}{Test Dataset $\Delta{\alpha}$ Reduction} \\ \cline{2-4}
    & D1 & D2 & D3 \\ \hline
    LSTM & 92.29\% & 92.11\% & 91.81\% \\ \hline
    Transformer & 90.91\% & 91.12\% & 90.53\% \\ \hline\hline
    \end{tabularx}
    \label{tb4}
\end{table}


LSTM and Transformer perform exceptionally well in reducing the final integration error of the IMU, denoted as $\Delta{\alpha}$, across all three test sets, with errors reduced by over 90\% in Table {\ref{tb4}}. 

\subsection{Analysis of Window Dimensions}

For the LSTM, we adjust the model's input data window from the original 32-step continuous sequence (16s) to 20 steps (10s), 10 steps (5s), and 6 steps (3s) to evaluate whether the model can effectively predict IMU data biases within shorter time frames. For the Transformer, given that the original model's input window dimension is 100 steps, we add a test with a 32-step input window to assess its performance under a shorter input sequence.

\begin{table}[]
    \caption{LSTM's Loss Across Different Input Window\label{tab:performance}}
    \centering
    \begin{tabularx}{\columnwidth}{|>{\centering\arraybackslash}X|>{\centering\arraybackslash}X|>{\centering\arraybackslash}X|>{\centering\arraybackslash}X|}\hline\hline
    \multirow{2}{*}{Input Window} & \multicolumn{3}{c|}{Test Dataset Training Loss} \\ \cline{2-4}
    & D1 & D2 & D3 \\ \hline
    20 & 0.021 & 0.020 & 0.011 \\ 
    10 & 0.025 & 0.025 & 0.013 \\ 
    6 & 0.024 & 0.033 & 0.011 \\ \hline\hline
    \end{tabularx}
    \label{tb5}
    \vspace{-0.3cm}
\end{table}

\begin{table}[]
    \caption{Transformer's Loss Across Different Input Window\label{tab:performance}}
    \centering
    \begin{tabularx}{\columnwidth}{|>{\centering\arraybackslash}X|>{\centering\arraybackslash}X|>{\centering\arraybackslash}X|>{\centering\arraybackslash}X|}\hline\hline
    \multirow{2}{*}{Input Window} & \multicolumn{3}{c|}{Test Dataset Training Loss} \\ \cline{2-4}
    & D1 & D2 & D3 \\ \hline
    32 & 0.021 & 0.008 & 0.013 \\ 
    20 & 0.018 & 0.006 & 0.011 \\ 
    10 & 0.020 & 0.007 & 0.012 \\ 
    6 & 0.029 & 0.014 & 0.014 \\ \hline\hline
    \end{tabularx}
    \label{tb6}
    \vspace{-0.3cm}
\end{table}

Table \ref{tb5} and Table \ref{tb6} show the model's performance on the test set with different IMU data input window dimensions. We find that as the input window decreases, both the LSTM and Transformer can still achieve effective convergence, indicating that both models are capable of quickly learning human movement patterns and providing reasonable predictions even within short time frames.



\subsection{Analysis of Models' Generalization Capability Across Different Motion Styles}

We train both two models using only one of the three motion styles as the training set and test it on the validation sets of the other two motion styles. This approach is used to assess whether the model could learn general patterns of human movement from a single type of action.

\begin{table}[h]
    \caption{LSTM Loss Value Across Different Movements\label{tab:LSTM diff move}}
    \centering
    \begin{tabularx}{\columnwidth}{|>{\centering\arraybackslash}X|>{\centering\arraybackslash}X|>{\centering\arraybackslash}X|}\hline\hline
    Training Movement & Test Movement A & Test Movement B \\ \hline
    Walk & 0.067(Run) & 0.033(Stair) \\ 
    Run & 0.021(Walk) & 0.012(Stair) \\ 
    Stair & 0.045(Walk) & 0.040(Run) \\ \hline\hline
    \end{tabularx}
    \label{tb8}
    \vspace{-0.5cm}
\end{table}

\begin{table}[h]
    \caption{Transformer Loss Value Across Different Movements\label{tab:Trans diff move}}
    \centering
    \begin{tabularx}{\columnwidth}{|>{\centering\arraybackslash}X|>{\centering\arraybackslash}X|>{\centering\arraybackslash}X|}\hline\hline
    Training Movement & Test Movement A & Test Movement B \\ \hline
    Walk & 0.014(Run) & 0.024(Stair) \\ 
    Run & 0.035(Walk) & 0.021(Stair) \\ 
    Stair & 0.057(Walk) & 0.029(Run) \\ \hline\hline
    \end{tabularx}
    \label{tb9}
    \vspace{-0.5cm}
\end{table}

As shown in Table \ref{tb8} and Table \ref{tb9}, both two models demonstrate good convergence on the other two movement patterns even when trained on only a single movement pattern. This validates the models' generalization capability across multiple movement patterns, Indicating that both models can be effectively applied to calibrate different movement patterns, even when they are trained on just one type of movement.

\subsection{Analysis of Model Generalizability Across Different IMUs}

After evaluating the model with the IMU in the Livox Mid 360, we further test its cross-sensor generalization capabilities using the VectorNav IMU. Building on previous findings, we consistently use a sequence length of six in a 6-step input window to ensure that the model performs reliably across various sensor data conditions. Based on the trained model, we apply it on the validation sets, and the results are shown in Table \ref{tb11}.

\begin{table}[h]
    \caption{Models' Loss On VectorNav IMU\label{tab:VEC IMU LOSS}}
    \centering
    \begin{tabularx}{\columnwidth}{|>{\centering\arraybackslash}X|>{\centering\arraybackslash}X|>{\centering\arraybackslash}X|>{\centering\arraybackslash}X|}\hline\hline
    \multirow{2}{*}{Model} & \multicolumn{3}{c|}{Test Dataset Loss} \\ \cline{2-4}
    & D1 & D2 & D3 \\ \hline
    LSTM & 0.077 & 0.092 & 0.093 \\ \hline
    Transformer & 0.088 & 0.065 & 0.060 \\ \hline\hline
    \end{tabularx}
    \label{tb10}
    \vspace{-0.3cm}
\end{table}

\begin{table}[h]
    \caption{$\Delta{\alpha}$ Reduction On VectorNav IMU\label{tab:delta alpha vec imu}}
    \centering
    \begin{tabularx}{\columnwidth}{|>{\centering\arraybackslash}X|>{\centering\arraybackslash}X|>{\centering\arraybackslash}X|>{\centering\arraybackslash}X|}\hline\hline
    \multirow{2}{*}{Model} & \multicolumn{3}{c|}{Test Dataset $\Delta{\alpha}$ Reduction} \\ \cline{2-4}
    & D1 & D2 & D3 \\ \hline
    LSTM & 60.22\% & 54.80\% & 53.77\% \\ \hline
    Transformer & 59.72\% & 52.55\% & 53.32\% \\ \hline\hline
    \end{tabularx}
    \label{tb11}
    \vspace{-0.3cm}
\end{table}


As shown in Table \ref{tb11}, the cross-sensor generalization capabilities of the two models were validated. Even when tested with data from different IMU sensors, the models still achieved an improvement of over 50\% in error reduction.
Although the performance on the VectorNav IMU shows a slight decline compared to the initial sensor, the overall performance and robustness of the models are effectively validated. This indicates that the models possess a certain level of adaptability across different hardware platforms and can provide relatively stable results in cross-device applications.





\section{Conclusion} \label{section_conclusion}

We present a novel head-mounted Inertial Measurement Unit (IMU) dataset with ground truth data, featuring contributions from ten participants engaged in various activities. This dataset addresses challenges inherent in current localization systems, which often struggle under adverse environmental conditions. Our investigation into advanced neural network methods, including Long Short-Term Memory (LSTM) and Transformer networks, demonstrates a remarkable 90\% improvement in IMU integration accuracy. These findings validate the effectiveness of these methods in correcting IMU biases and enhancing positioning precision. We further assess the performance of these approaches across different IMU data window dimensions, motion patterns, and sensor types, confirming the viability of advanced neural network techniques for helmet-based localization. To benefit the SLAM society, we have made the dataset publicly available, complete with evaluation metrics and a baseline for future studies in this domain.


\bibliographystyle{ieeetr}
\bibliography{ref} 
\end{document}